\documentclass[letterpaper, 10 pt, conference]{ieeeconf}  % Comment this line out if you need a4paper

\IEEEoverridecommandlockouts                              % This command is only needed if
\usepackage{calrsfs}
\usepackage{graphicx}
\usepackage{gensymb}
\usepackage{booktabs}
\usepackage{multicol}
\usepackage{multirow}
\usepackage{longtable}
\usepackage{graphics} % for pdf, bitmapped graphics files
\usepackage{epsfig} % for postscript graphics files
\usepackage{times} % assumes new font selection scheme installed
\usepackage{amsmath} % assumes amsmath package installed
\usepackage{amssymb}  % assumes amsmath package installed
\usepackage{mathtools}
\usepackage{amsmath}
\usepackage{cite}
\usepackage{bm}
\usepackage{threeparttable}

\DeclareMathAlphabet{\pazocal}{OMS}{zplm}{m}{n}

\title{\LARGE \bf
Translation Invariant Global Estimation of Heading Angle \\Using Sinogram of LiDAR Point Cloud
}

\author{Xiaqing Ding\textsuperscript{1,2}, Xuecheng Xu\textsuperscript{1}, Sha Lu\textsuperscript{1}, Yanmei Jiao\textsuperscript{1}, Mengwen Tan\textsuperscript{2}, Rong Xiong\textsuperscript{1},\\ Huanjun Deng\textsuperscript{2}, Mingyang Li\textsuperscript{2}, Yue Wang\textsuperscript{1*} % <-this % stops a space

\thanks{*This work was supported by the National Nature Science Foundation of China under Grant 61903332, the National Key R\&D Program of China under Grant 2018YFB1600804 and ZheJiang Program in Innovation, Entrepreneurship and Leadership Team (2018R01017).}% <-this % stops a space
\thanks{*This work was supported by Alibaba Group through Alibaba Innovative Research Program.}
\thanks{$^1$ State Key Laboratory of Industrial Control Technology and Institute of Cyber-Systems and Control, Zhejiang University, Zhejiang, China. }
\thanks{$^2$ Alibaba Group, Hangzhou, 310052, China. }%
\thanks{$^*$ Corresponding author {\tt\small wangyue@iipc.zju.edu.cn}.}
}

\begin{document}

\maketitle
\thispagestyle{empty}
\pagestyle{empty}

% ground
%%%%%%%%%%%%%%%%%%%%%%%%%%%%%%%%%%%%%%%%%%%%%%%%%%%%%%%%%%%%%%%%%%%%%%%%%%%%%%%%
\begin{abstract}
Global point cloud registration is an essential module for localization, of which the main difficulty exists in estimating the rotation globally without initial value. With the aid of gravity alignment, the degree of freedom in point cloud registration could be reduced to 4DoF, in which only the heading angle is required for rotation estimation. In this paper, we propose a fast and accurate global heading angle estimation method for gravity-aligned point clouds. Our key idea is that we generate a translation invariant representation based on Radon Transform, allowing us to solve the decoupled heading angle globally with circular cross-correlation. Besides, for heading angle estimation between point clouds with different distributions, we implement this heading angle estimator as a differentiable module to train a feature extraction network end-to-end. The experimental results validate the effectiveness of the proposed method in heading angle estimation and show better performance compared with other methods.
%Global localization is essential for robot navigation, of which the first step is to retrieve a query from the map database. This problem is called place recognition. In recent years, LiDAR scan based place recognition has drawn attention as it is robust against the environmental change. In this paper, we propose a LiDAR-based place recognition method, named Differentiable Scan Context with Orientation (DiSCO), which simultaneously finds the scan at a similar place and estimates their relative orientation. The orientation can further be used as the initial value for the down-stream local optimal metric pose estimation, improving the pose estimation especially when a large orientation between the current scan and retrieved scan exists. Our key idea is to transform the feature learning into the frequency domain. We utilize the magnitude of spectrum as the place signature, which is theoretically rotation-invariant. In addition, based on the differentiable phase correlation, we can efficiently estimate the global optimal relative orientation using the spectrum. With such structural constraints, the network can be learned in an end-to-end manner, and the backbone is fully shared by the two tasks, achieving interpretability and light weight. Finally, DiSCO is validated on the NCLT and Oxford datasets with long-term outdoor conditions, showing better performance than the compared methods.
%% \footnote{Codes are released at https://github.com/MaverickPeter/DiSCO-pytorch.}
\end{abstract}

%%%%%%%%%%%%%%%%%%%%%%%%%%%%%%%%%%%%%%%%%%%%%%%%%%%%%%%%%%%%%%%%%%%%%%%%%%%%%%%%

\section{Introduction}

Global localization is important in many robotic applications. During recent years, lots of successful vision based methods are presented \cite{2015Visual,Arandjelovic2018,tang2020adversarial,ijrrrevisited}, following a staged process of place recognition then pose estimation. However, vision based methods are sensitive to strong appearance and viewpoint changes, which cannot be avoided in long-term outdoor applications. Alternatively, LiDAR based methods are much more stable to appearance variations and can perform place recognition when the robot revisits a place in a different viewpoint \cite{uy2018pointnetvlad,SCI2019,yin2018locnet,pcan2019,liu2019lpd}. 
% The 360-degree view of the LiDAR also improves the performance of place recognition when robot re-visits a place in different viewpoint\cite{yin2018locnet,pcan2019,liu2019lpd}. 
However, this advantage raises a new challenge for downstream pose estimation. Currently, the widely-used point cloud registration techniques e.g. ICP\cite{Pomerleau12comp}, have local convergence and call for good initial values. However, the heading angle between the retrieved laser scan and the current one might be large, leading to the failure of pose estimation. Some works are designed to solve point cloud registration globally by feature matching\cite{fpfh2009,dh3d2020,LeiTIP2017,TeaserTRO2021} or exploring the whole solution space under the branch-and-bound framework \cite{yang2013go,enqvist2009optimal}. But the former relies on the quality of the correspondences, while the latter is time-consuming.
% When the two scans have large difference in heading angle, these methods usually fail. 
Therefore, the global point cloud registration becomes the bottleneck to take advantage of the 360-degree view of the LiDAR, and solve the full global localization problem.

A more recent line of works takes advantage of the relatively planar terrain in many applications and IMU-aided coarse gravity alignment to reduce the pose estimation to 3 or 4DoF\cite{jiao20202,lu2019l3,2010Closed}, among which only the heading angle is nonlinear. Given that the retrieved point cloud is generally close to the current one, some works turn to focus on global heading angle estimation between two gravity-aligned point clouds\cite{oreos19,kim2018scan,xu2021disco,chen2020overlapnet}.
% As the place recognition can guarantee small translation between the two scans, this line of works focuses on global heading angle estimation. 
A popular way is to convert the point cloud representation from 3D Cartesian coordinate to range image so that the relative heading angle is reflected on the translational shift of the image, thus can be estimated by traversing all possible shifted images\cite{chen2020overlapnet} or direct regressed based on deep neural network\cite{oreos19}.
%The early leading work of heading estimation employs a deep neural network for direct regression \cite{xxx}. Following works improves the regression by introducing inductive bias in the intermediate representation. In \cite{overlapnet}, the spherical representation is used to represent angle difference in column shifting.
The weakness of this representation is the sensitivity to the quality of gravity alignment and the translation distance between point clouds. In \cite{xu2021disco,kim2018scan}, the polar bird-eye view (BEV) of point cloud is proposed to achieve the transformation from rotational difference to translational shift, which increases the tolerance to noise in pitch and roll. However, in this representation, the translation still couples with rotation and deteriorates the accuracy.

\begin{figure}
    \centering
    \includegraphics[width=0.49\textwidth]{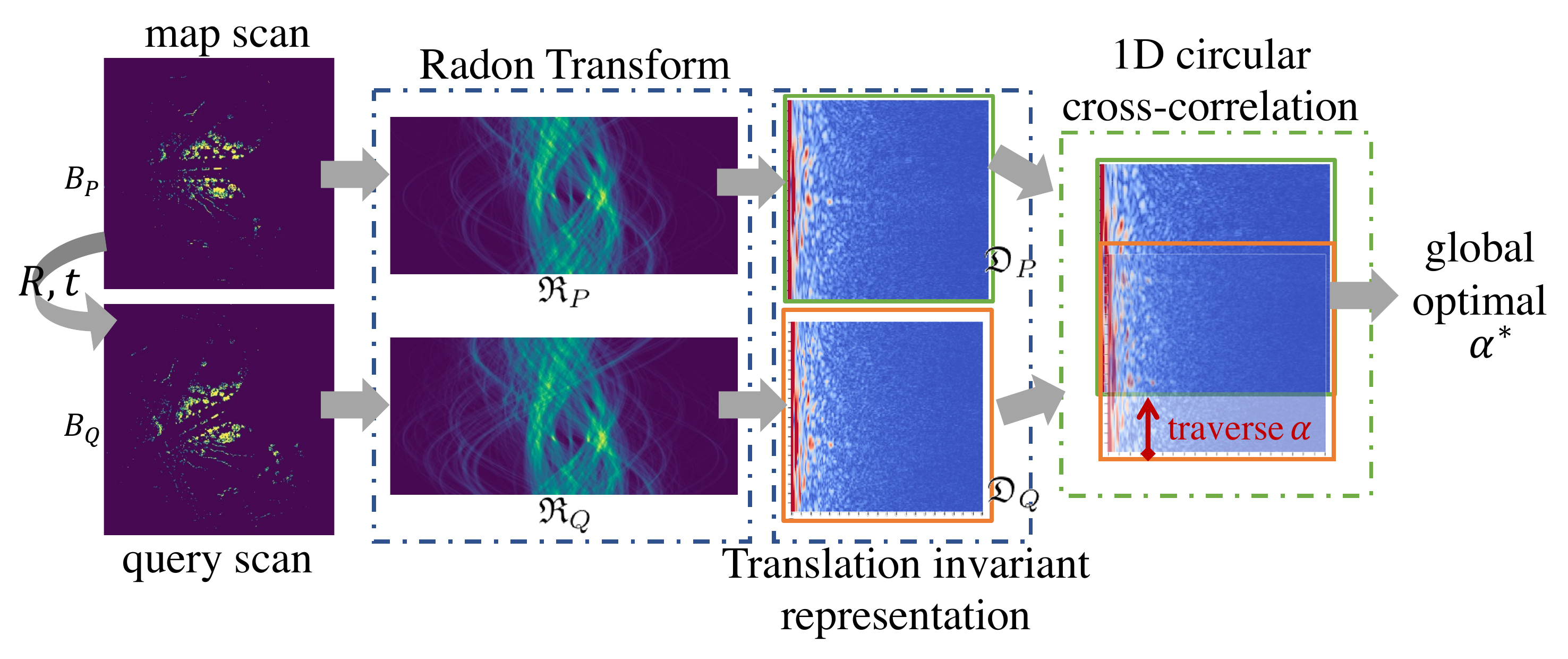}
    \caption{The pipeline of the global optimal heading angle estimator proposed in this paper.}
    \label{fig:framework}
\end{figure}

In this paper, we set to propose a translation invariant representation for global heading angle estimation between gravity-aligned point clouds. As shown in Fig. \ref{fig:framework}, the translation invariant representation is developed based on the sinogram generated by the Radon Transform (RT), 
% In this paper, we set to address the global estimation of heading angle by proposing a sinogram representation. The sinogram is the target domain of the Radon transform (RT), 
which is widely used in medical computed tomography \cite{BARRETT1984217,1991Mathematical,2004Nonlinear}. The sinogram is parameterized by integrations along a set of scanning lines, where the x- and y-axis refer to the slope and offset of the scanning lines, respectively. 
Under this representation, the heading angle between point clouds is reflected as the shift along the slope-related axis, while the translation distance is reflected as the shift along the offset-related axis coupled with the slope. We further eliminate the shift along the offset-related axis, making the resultant representation invariant with translation, and can be efficiently solved based on the circular cross-correlation along the slope-related axis.
% By applying RT to the LiDAR point cloud BEV, we have its sinogram encoding the one-dimension translation and heading. Then we propose a global solver for the shift between the sinograms of the two LiDAR point clouds, arriving at the heading angle. 
Experiments show that our method preserves the robustness to gravity noise because of BEV \cite{xu2021disco}, and translation invariance because of sinogram. As there exist fast algorithms to implement RT, the onboard processing frame rate can exceed 500Hz. In addition, we only use the binary occupancy information to generate the BEV, thus the storage can be largely saved, which is desired by light-weighted autonomous robots. Furthermore, we propose to integrate RT as a differentiable heading angle estimation module in a deep network that extracts dense features from BEVs. In this way, the heading angle estimation between point clouds in different distributions, e.g., scan and submap, can also be estimated with high accuracy, which releases the storage pressure of keeping all raw scans when building the map. Finally, toycase studies and experimental results both validate the superior performance of the sinogram based method in effectiveness and efficiency. In summary, the contribution consists of
\begin{itemize}
\item A sinogram based heading angle estimation method, which achieves translation invariance and global convergence.
% \item A differentiable RT is embedded in a deep neural network for dense feature extraction, enabling heading angle estimation between scan and submap.
\item A differentiable RT, embedded in a deep neural network for dense feature extraction, enabling heading angle estimation between scan and submap.
\item Toycase and real-world experimental results verify the theoretical insights and demonstrate better performance in testing and generalization trials.
\end{itemize}

\section{Related Work}

\subsection{Place recognition}

Place recognition is the first stage of global localization. 
% It finds a scan, or a submap taken at the similar place with the current one. 
As the in-place rotation does not change the place, this module generally only cares about the translation error between the query place and the retrieved place. 
% This stage is not the focus of this paper, so we briefly introduce the related works for completeness of application purpose. 
% The early success of image based place recognition is based on bag-of-words (BoW) descriptor \cite{GalvezTRO12}, which is still popular in modern visual SLAM systems \cite{mur2017orb,qin2018vins}. Since BoW is built upon the keypoint feature, it is not robust to appearance change. To overcome the problem, dense descriptor is proposed to relieve the problem such as the popular NetVLAD \cite{Arandjelovic2018,torii201524,tang2020adversarial}.
% However, such methods achieve robustness by training on massive diverse dataset.
% Alternatively, 
LiDAR has gained lots of attention in recent years due to its robustness to appearance changes in the sensor data level. Early trials extend the visual place recognition\cite{Arandjelovic2018} to LiDAR, such as PointNetVLAD \cite{uy2018pointnetvlad}. Since the geometry in LiDAR is fully observed, more studies try to utilize this property for special architecture design for LiDAR and achieve better performance\cite{kim2018scan,xu2021disco,yin2018locnet}. With the 360-degree view, point cloud based place recognition can retrieve the correct place when the current visit is in the opposite direction of the mapping visit \cite{seqlpd19,liu2019lpd,coral21}.

% \subsection{Feature based pose estimation ? 3D point cloud Registration?}
\subsection{3D point cloud registration}
% DD TODO
% feature svd e2e learning
% feature learning, robust estimator

6DoF pose estimation is the second stage of global localization following place recognition. Generally, LiDAR based pose estimation is achieved by 3D point cloud registration in local or global optimal ways. Local solutions \cite{segal2009generalized,Pomerleau12comp,NDT2003} construct data association across two point clouds based on nearest neighbor searching in Cartesian space, thus an initial pose is required for accurate pose estimation. Global solutions can be divided into feature matching based approaches and correspondence-free approaches. Feature matching based methods extract handcrafted or learned keypoint features and construct data association based on the descriptors\cite{fpfh2009,LeiTIP2017,3DFeatNet,dh3d2020}. As it's difficult to guarantee the reliability of point cloud features, RANSAC\cite{1987Random,aiger20084,fpfh2009} or M-estimation\cite{zhou2016fast,mactavish2015all} are widely applied to suppress
restricted outlier matches. To improve the global convergence and the robustness against high outlier ratio, TEASER\cite{TeaserTRO2021} designs invariant measurements that remove outliers before feeding correspondences into backend solver, and achieves robustness to more than 90$\%$ outliers. However, the optimality is still coupled with the quality of matches. Correspondence-free approaches are designed based on the branch-and-bound framework, in which the solution space is divided and explored for global optimal estimation\cite{yang2013go,enqvist2009optimal}. Though with optimality guarantee, these approaches are extremely time-consuming and are not suitable for real-time applications.

\subsection{Heading angle estimation}

% Another line of pose estimation return to the dimension of the original problem.
Considering the relatively flat local terrain, IMU aided coarse gravity alignment, and the fully observed geometry, the pose estimation can be reduced to {3DoF} in many applications, e.g., autonomous driving. Furthermore, as the correct place recognition can guarantee the adjacency between two point clouds, recent methods focus on the global estimation of the heading angle. \cite{gkim-2018-iros} represents the place into scan context, and estimates the heading angles by column shift, which takes advantage of the cyclic correlation for holistic alignment. \cite{oreos19} transforms the 3D point cloud into the spherical coordinate based range image, and estimates the heading angle by direct regression using a neural network. As there is no explicit architecture constraint in the estimator, the accuracy and generalization are limited. To overcome this problem, inductive bias is introduced in the network architecture design. In \cite{chen2020overlapnet} the cyclic correlation is utilized to estimate the heading angles between features learned from range images. This representation could demonstrate good performance in planar motion. However, this representation is sensitive to the gravity alignment and the relative translation. To address the first problem, BEV representation is proposed, and frequency transform is applied for heading angle estimation in \cite{xu2021disco,2020Dpcn}. These methods even show good performance on heterogeneous sensor modalities. However, the translation is still coupled in the representation, which biases the estimation results. In this paper, we set to equip the heading angle estimator with certifiable translation invariance.
% ? DPCN is not vulnerable to translation. Not mentioned? 

%\begin{figure*}[t]
%\centering
%\includegraphics[scale=0.5]{figures/Overview.png}
%\caption{Overview of our proposed framework DiSCO. The data representation can be any polar BEV image.}
%\label{fig:Pipeline}
%\vspace{-0.3cm}
%\end{figure*}

\section{Preliminaries on Sinogram}

\begin{figure}
    \centering
    \includegraphics[width=0.46\textwidth]{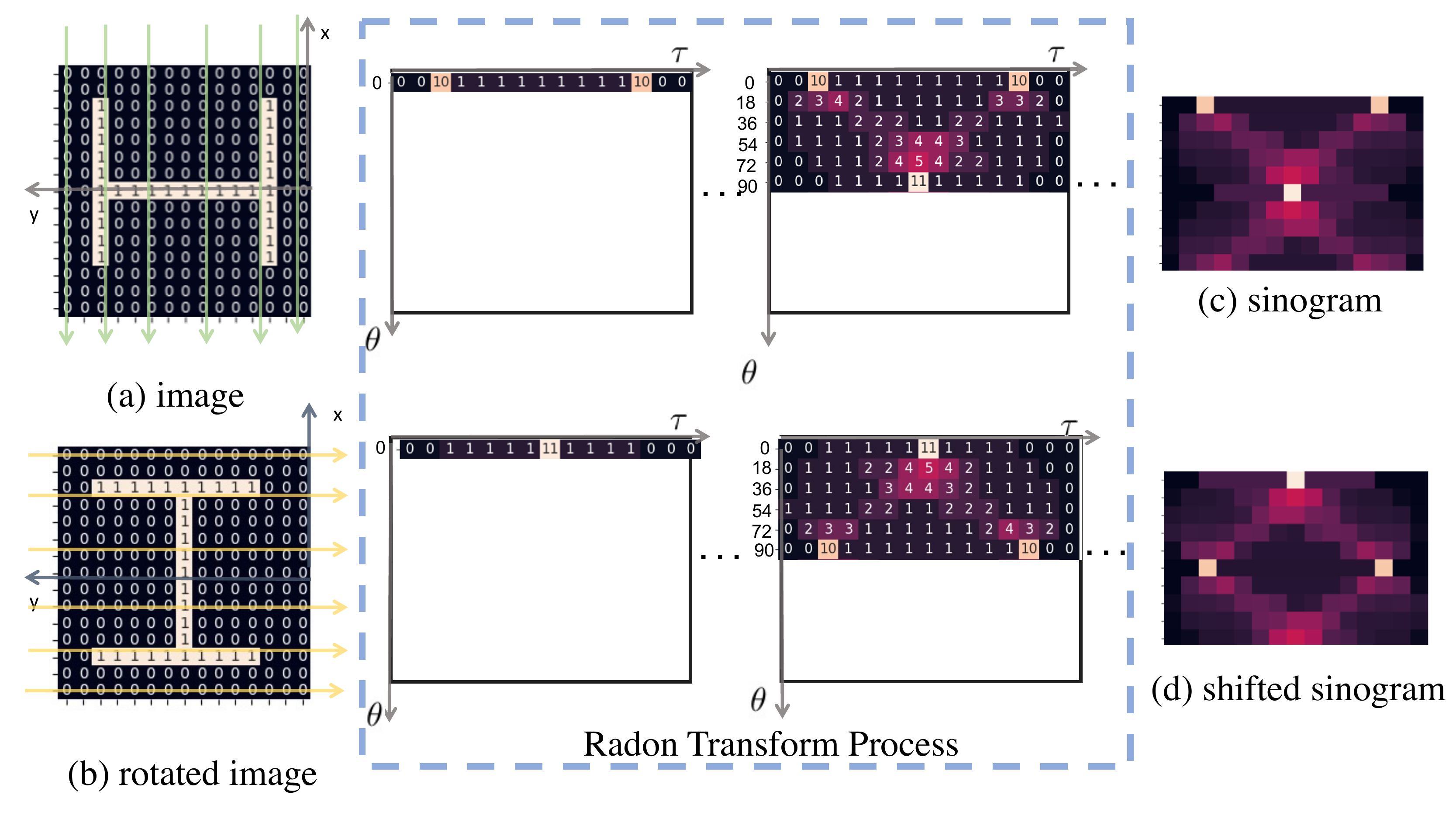}
    \caption{Radon Transform process and the shift property. The image in (b) is generated by rotating the image in (a) by 90$\degree$, and their corresponding sinograms are drawn in (c) and (d), in which the rotated angle can be inferred by the shift along the y-axis in (c) and (d). The dashed box shows some intermediate results at $0\degree$ and $90\degree$ incident angles. 
    % The image in (b) is generated by rotating the image in (a) by 90$\degree$, which can be inferred by the shift along the y-axis in (c) and (d).
    }
    \label{fig:rt}
\end{figure}

We first briefly introduce the RT theory. Given an image $I$ with size $S\times S$, RT is a linear integral transform that reformulates the image with line parameters.
% which takes image on plane to space parameterized by scanning line on the plane.
Specifically, given a parameterized scanning line $k u+\tau =0$, where $k$ is the slope controlled by the incident angle $\theta$, i.e., $k(\theta)=[sin(\theta), cos(\theta)]$, $\tau$ is the offset, and $u\in [0,S)\times[0,S)$ is the pixel position, we can calculate the integral along this line on the image as
\begin{equation}\label{lineint}
  \mathfrak{R}(\theta,\tau) \triangleq \int I(u) du,~~~~k(\theta)u+\tau = 0
\end{equation}
By screening the space of $\{\theta\}$ and $\{\tau\}$, i.e., the whole space of the scanning line parameters on the image, we finish RT of the image $I$, and achieve a completed $\mathfrak{R}$ called the sinogram of $I$. Note that the sinogram is also a 2D polar image with $\theta$ and $\tau$ spanned along the two axes. Different from the polar representation in \cite{oreos19,chen2020overlapnet,kim2018scan,xu2021disco}, the rotation angle and its reference coordinate are controlled externally instead of being anchored with the robot positions. The process can be regarded as rotating a scanner for integration with an array of parallel scanning lines as shown in Fig. \ref{fig:rt}.

\textbf{Shift property of sinogram:}
We now derive the relationship between the 2D transformation on image space and the corresponding shift on the sinogram. Let $\{R(\alpha), t\}$ denotes the 2D transformation, in which $R(\alpha) \in \mathbb{SO}(2)$ is the rotation matrix constructed by angle $\alpha$, and $t\in \mathbb{R}^2$ is the translation vector. Each pixel on the transformed image $I'$ can be derived as
\begin{equation}\label{transform_image}
    I'(u)=I(R(\alpha)u+t)
\end{equation}
Thus for any scanning line $k(\theta)u+\tau=0$ on $I$, we can find a scanning line $k(\theta')u+\tau'=0$ that integrates out the same result on $I'(u)$, of which the slope $k(\theta')=R(\alpha)k(\theta)=k(\theta+\alpha)$, and the offset $\tau'=k(\theta)t+\tau$. In this way, the corresponding sinogram can be derived as
\begin{equation}\label{transform_sino}
    \mathfrak{R}'(\theta, \tau)=\mathfrak{R}(\theta-\alpha, \tau-k(\theta)t)
\end{equation}
From Eq(\ref{transform_sino}) we can notice that under the RT based representation, the rotational and translational parts of 2D transformation are decoupled, and are reflected by shifts along the two axes of the sinogram, respectively. With this important shift property, we can build an estimator of heading angle in the next section.

% Thus different from scan context and spherical representations in \cite{oreos19,chen2020overlapnet}, the rotation between

% 旋转相关的维度不受平移影响，

% When we apply a rotational transformation with angle $\theta_d \in [0,\pi)$ centering at the origin to the image, we have a new image $I_{\theta_d,0}(u)$. Then the sinograms of $I(u)$ and $I_{\theta_d,0}(u)$ are related by a shift only in the axis of $\theta$ as:
% \begin{equation}\label{rtrot}
%   \mathfrak{R}_{\theta_d,0}(\theta,b) = \mathfrak{R}(\theta-\theta_d,b)
% \end{equation}
% It is obvious that we can rotate the array of scan line with $\theta_d$ to keep two results the same. When we apply a translation $d\in \mathbb{R}^2$ to generate an new image $I_{0,d}(u)$, we have
% \begin{equation}\label{rttrans}
%   \mathfrak{R}_{0,d}(\theta,b) = \mathfrak{R}(\theta,b+n^T d)
% \end{equation}
% The proof can be found in \cite{https://backend.orbit.dtu.dk/ws/portalfiles/portal/5529668/Binder1.pdf}. One can see that translation of an image leads to shift only in the axis of distance $b$ in the sinogram, but the length of shift is not uniform across lines with different $\theta$. With the two important shift properties of RT, we are able to build an estimator of heading angle in the next section.

\section{Global Estimation of Heading Angle}\label{base heading}

\subsection{Problem statement}

Given a query 3D point cloud $Q \triangleq \{q\}$ and a point cloud $P \triangleq \{p\}$ retrieved by place recognition, global registration aims at estimating the 3D rigid transform between the two point clouds for localization. Following a common fact that the robot moves on locally planar terrain and the gravity could be aligned by IMU, we consider that the relative pitch and roll are slight, thus the nonlinearity for global registration only relates to the heading angle.
% reducing the 3D rigid transform to 2D, i.e., rotation matrix $R(\alpha)$ and translation vector $t$.

If translation disturbance is neglected, the global optimal heading angle could be solved by traversing all rotational angles \cite{chen2020overlapnet, xu2021disco}. However, the representations in these methods are not translation invariant. When the translation can not be neglected, the estimated optimal results would be biased. We set to consider the effect of translation, and still solve the heading angle in global.

We follow \cite{xu2021disco} to generate BEV images of $Q$ and $P$, namely $B_Q$ and $B_P$, which suppress the influence of the noisy gravity alignment. To get rid of the effect from the absolute heights of different robots, we encode the BEV with the only binary occupancy information, which also largely saves the storage. At this point, by regarding the small components as noise $n_B$, the relative 2D transformation between the two BEV images leads to
%we reduce the 3D rigid transform to 2D, i.e., rotation matrix $R$ and translation vector $t$, leading to:
\begin{equation}\label{imagealign}
  B_Q(u) = B_P(R(\alpha)u+t) + n_B
\end{equation}
where $u$ is the position vector indexed in the BEV. Now the problem of estimating the relative heading angle could be defined as
\begin{equation}\label{problemstate}
  \alpha^* = \arg \min_{\alpha} \|B_Q(u) - B_P(R(\alpha)u+t)\|
\end{equation}
Unfortunately, it is difficult to solve it directly to find the global optimal value.
% The difficulty to this problem is the disturbance of translation $t$. To eliminate the disturbance, methods in \cite{overloopnet,disco} simply neglects this term, which introduces error to the heading angle estimation.
% If neglecting the disturbance of translation, this problem could be solved by traversing all rotational angles by resolution.
% We set to consider the effect of $t$, and still solve $\alpha^*$ in global without referring to an initial value.

\subsection{Global sinogram shift solver}
% By encoding the point cloud as BEV representation, $\alpha$ and $t$ are reflected by the image rotation and translation between $B_Q$ and $B_P$. In fact, translation $t$ and image translation are related by a resolution. As the resolution is determined manually, we also use $t$ as image translation for notation clarity.
We apply RT to both BEVs, leading to $\mathfrak{R}_Q$ and $\mathfrak{R}_P$. Referring to (\ref{transform_sino}), we have
\begin{equation}\label{rtbqbp}
  \mathfrak{R}_Q(\theta,\tau) = \mathfrak{R}_P(\theta-\alpha^*,\tau-k(\theta)t)+n_R
\end{equation}
where $n_R$ denotes the integrated noise. To solve $\alpha$ using sinogram, we have to eliminate the shift coupled with incident angle and translation along the axis of $\tau$.

\textbf{Shift invariant transformation:} Our idea is to transform the row vectors on the sinogram into a shift invariant representation, so that the effect from $t$ is eliminated and only row shift exists between two sinograms. We achieve this by applying 1D discrete Fourier transform (DFT) on each row vector and calculate the magnitude of the result
% Our idea to eliminate the row-wise shift on the sinogram is to build a shift invariant representation for each $\theta$, so that the non-uniform shifts problem is avoided.
% Given a specific $\theta$, say $\tilde{\theta}$, we have a 1D row of sinogram denoted as $\mathfrak{R}_{Q,\tilde{\theta}}(b)$. Applying 1D discrete Fourier transform (DFT) to the row, we have the frequency spectrum, of which the magnitude is shift invariant:
% \begin{equation}\label{rtbqbpdft}
%   |\mathfrak{F}(\mathfrak{R}_{Q,\tilde{\theta}}(b))| = |\mathfrak{F}(\mathfrak{R}_{Q,\tilde{\theta}}(b+b'))|
% \end{equation}
\begin{equation}\label{rtbqbpdft}
\begin{split}
      |\mathfrak{F}(\mathfrak{R}_{Q}(\theta_i,\tau))| &= |\mathfrak{F}(\mathfrak{R}_{P}(\theta_i-\alpha^*,\tau-k(\theta_i)t)))|\\
      &=|\mathfrak{F}(\mathfrak{R}_{P}(\theta_i-\alpha^*,\tau)))|
\end{split}
\end{equation}
where $\mathfrak{F}(\cdot)$ is the DFT operator and $\theta_i$ denotes $i$th row on the sinogram. By stacking the magnitude of the spectrum for each row on $\mathfrak{R}_Q$ and $\mathfrak{R}_P$, we can transform the two sinograms into new images $\mathfrak{D}_Q$ and $\mathfrak{D}_P$ as shown in Fig.\ref{fig:framework} that hold
\begin{equation}\label{magnitude}
    \mathfrak{D}_Q(\theta,\tau)=\mathfrak{D}_P(\theta-\alpha^*,\tau)+n_D
\end{equation}
where $n_D$ denotes the noise on the magnitude images.
% \begin{equation}\label{Dq}
%   \mathfrak{D}_Q(\theta) \triangleq \left[
%                      \begin{array}{c}
%                       |\mathfrak{F}(\mathfrak{R}_{Q,\theta_0}(b))| \\
%                       \vdots \\
%                       |\mathfrak{F}(\mathfrak{R}_{Q,\theta_{K-1}}(b))| \\
%                      \end{array}
%                   \right]
% \end{equation}
% where $K$ is the resolution for screening space of $\theta$. Correspondingly, we have
% \begin{equation}\label{Dp}
% \setlength{\arraycolsep}{0.6pt}
%   \mathfrak{D}_P(\theta) \triangleq \left[
%                      \begin{matrix}
%                      \setlength{\arraycolsep}{0.6pt}
%                      \begin{smallmatrix}
%                       |\mathfrak{F}(\mathfrak{R}_{P,\theta_0}(b+n^T t))| \\
%                       \vdots \\
%                       |\mathfrak{F}(\mathfrak{R}_{P,\theta_{K-1}}(b+n^T t))| \\
%                       \end{smallmatrix}
%                      \end{matrix}
%                   \right] = \left[
%                      \begin{matrix}
%                      \setlength{\arraycolsep}{0.6pt}
%                      \begin{smallmatrix}
%                       |\mathfrak{F}(\mathfrak{R}_{P,\theta_0}(b))| \\
%                       \vdots \\
%                       |\mathfrak{F}(\mathfrak{R}_{P,\theta_{K-1}}(b))| \\
%                       \end{smallmatrix}
%                      \end{matrix}
%                   \right]
% %
% \end{equation}
% where the equality holds according to (\ref{rtbqbpdft}).

\textbf{Estimation of heading angle:}
% Based on (\ref{rtbqbp}) and (\ref{magnitude}), by shifting the rows in $\mathfrak{D}_P(\theta)$ for $\alpha^*$, we have the core result of this paper:
% \begin{equation}\label{dqdp}
%   \mathfrak{D}_Q(\theta) = \mathfrak{D}_P(\theta-\alpha^*)+n
% \end{equation}
The global optimal result $\alpha^*$ can now be estimated by traversing all the possible $\alpha$ and calculating the correlation between $\mathfrak{D}_Q$ and all the row-wise shifted $\mathfrak{D}_P$
\begin{equation}\label{opt}
  \alpha^* = \arg\max_{\alpha} \sum_{\theta_i} \mathfrak{D}_Q(\theta_i,\tau)\mathfrak{D}_P(\theta_i-\alpha,\tau)^T
\end{equation}
% \begin{equation}\label{opt}
%   \alpha^* = \arg\max_{\alpha} \mathfrak{F}^{-1}(tr \{\mathfrak{F}_{\theta}(\mathfrak{D}_Q(\theta,\tau))\mathfrak{F}_{\theta}(\mathfrak{D}_P(\theta,\tau))^T\})
% \end{equation}
And this circular cross-correlation could be efficiently carried out by element-wise multiplication in the frequency space\cite{oppenheim1999discrete}. We define the circular cross-correlation process as $\mathfrak{C}(\mathfrak{D}_Q,\mathfrak{D}_P)$, of which the output is a $N$ dimensional vector with $N$ as the number of possible values of $\alpha$.
% Inversely, to estimate $\alpha^*$, we propose a correlation based solver for estimating the 1D shift between two matrices in global:

% where $\mathfrak{F}^{-1}(\cdot)$ is the iDFT, $tr\{\cdot\}$ is the trace operator and $\mathfrak{F}_{\theta}(\cdot)$ denotes that the 1D DFT is taken along the $\theta$ axis.

To summarize, the proposed estimator eliminates the effect of translation in two steps: 1) transforms translation as shift only in the axis of $\tau$ using RT, and 2) transforms all rows of sinogram into shift invariant magnitude of the frequency spectrum. In this way, the translation is not simply neglected but carefully isolated, which is expected to improve the accuracy of the heading angle estimation. In addition, since RT and DFT are all classical arithmetic operators, there are efficient implementations available, which makes onboard processing possible.

\section{Deep Feature Extraction for Sinogram}

\begin{figure}
    \centering
    \includegraphics[width=0.48\textwidth]{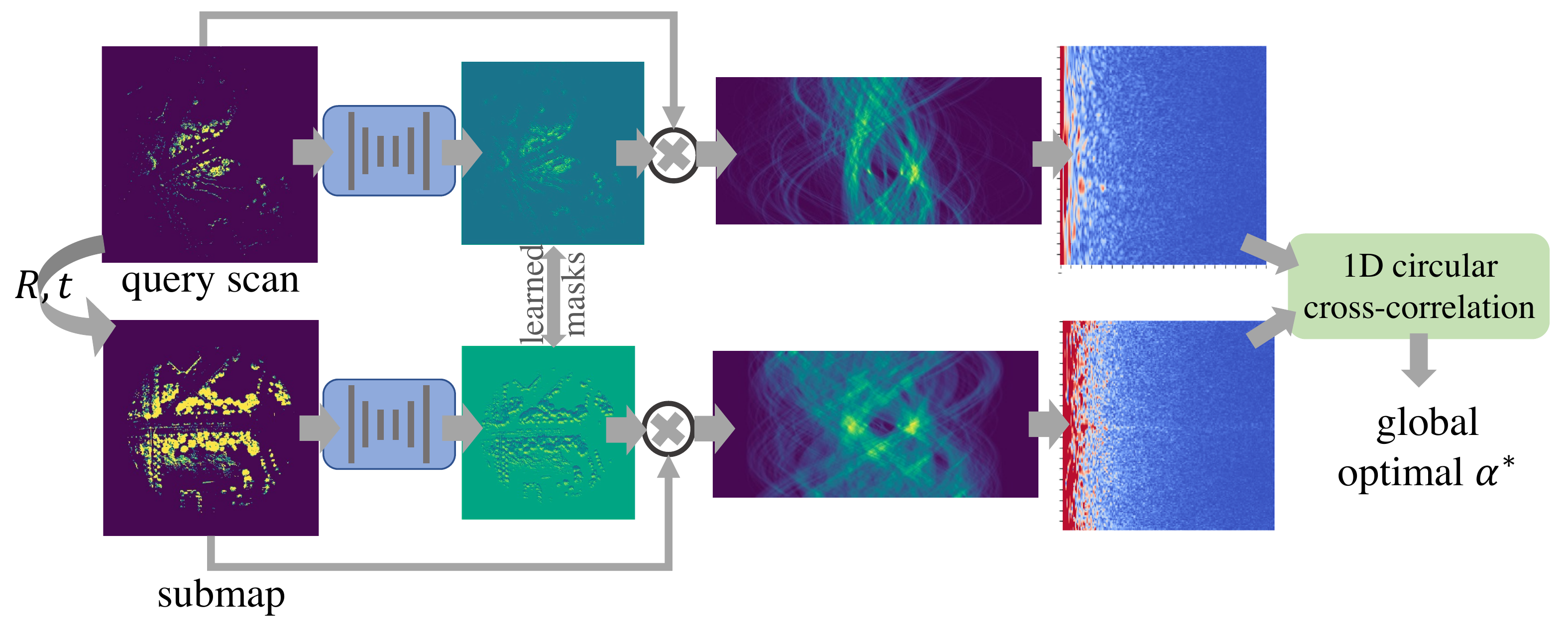}
    \caption{The pipeline of the global optimal heading angle estimator for point clouds in different distributions.}
    \label{fig:framework-dradon}
\end{figure}

When the query and retrieved point clouds are in the same distributions, their heading angle can be solved efficiently and globally following the process in Section \ref{base heading}. However, if the distributions are different, the correlation between sinograms would be biased, leading to wrong heading angle estimation. Here we utilize a network for feature extraction and integrate a differentiable RT based heading estimator to enable end-to-end training.
\subsection{Network architecture}
We apply two encoder-decoder networks in the UNet\cite{UNet} architecture for feature extraction on $\mathfrak{B}_Q$ and $\mathfrak{B}_P$. Then the heading angle is calculated by the process in Section \ref{base heading} masking the BEV images with the extracted feature maps as shown in Fig. \ref{fig:framework-dradon}. Note that the two networks do not share the weights, which are expected to learn the map from the different distributions of scan and submap.

\subsection{Loss function}
To make the RT based solver differentiable, we change the max operator in Eq (\ref{opt}) with the softmax operator to evaluate the probability distribution of heading angle
% \begin{equation}
%     p(\theta)=softmax(\mathfrak{F}^{-1}(tr \{\mathfrak{F}_{\theta}(\mathfrak{D}_Q(\theta,\tau))\mathfrak{F}_{\theta}(\mathfrak{D}_P(\theta,\tau))^T\}))
% \end{equation}
\begin{equation}
    p(\theta)=softmax(\mathfrak{C}(\mathfrak{D}_Q,\mathfrak{D}_P))
\end{equation}
The loss is designed as the Kullback-Leibler divergence (KLD) between $p(\theta)$ and the one-hot distribution $\bm{1}(\theta-\alpha^*)$ generated based on the groundtruth of heading angle $\alpha^*$.
\begin{equation}
    loss=KLD(p(\theta), \bm{1}(\theta-\alpha^*))
\end{equation}

\section{Experiments}
In this section, we first utilize a toycase study to validate the translation invariance property of the RT based representation and its superiority compared with the other polar image representations. Then we utilize two real-world datasets to demonstrate the efficiency and accuracy of the proposed RT based heading angle estimator, and its ability to deal with point clouds in different distributions.

\subsection{Implementation and Training}
During training the BEV image is generated with the resolution of $400\times400$, and one image pixel relates to $0.5$ meter. The network is implemented with PyTorch \cite{NEURIPS2019_9015} and the Adam \cite{kingma2014adam} optimizer is utilized for training with learning rate of $1e-4$. We randomly rotate the input BEV images for data augmentation.

\subsection{Toycase Study}
\begin{figure}
    \centering
    \includegraphics[width=0.38\textwidth]{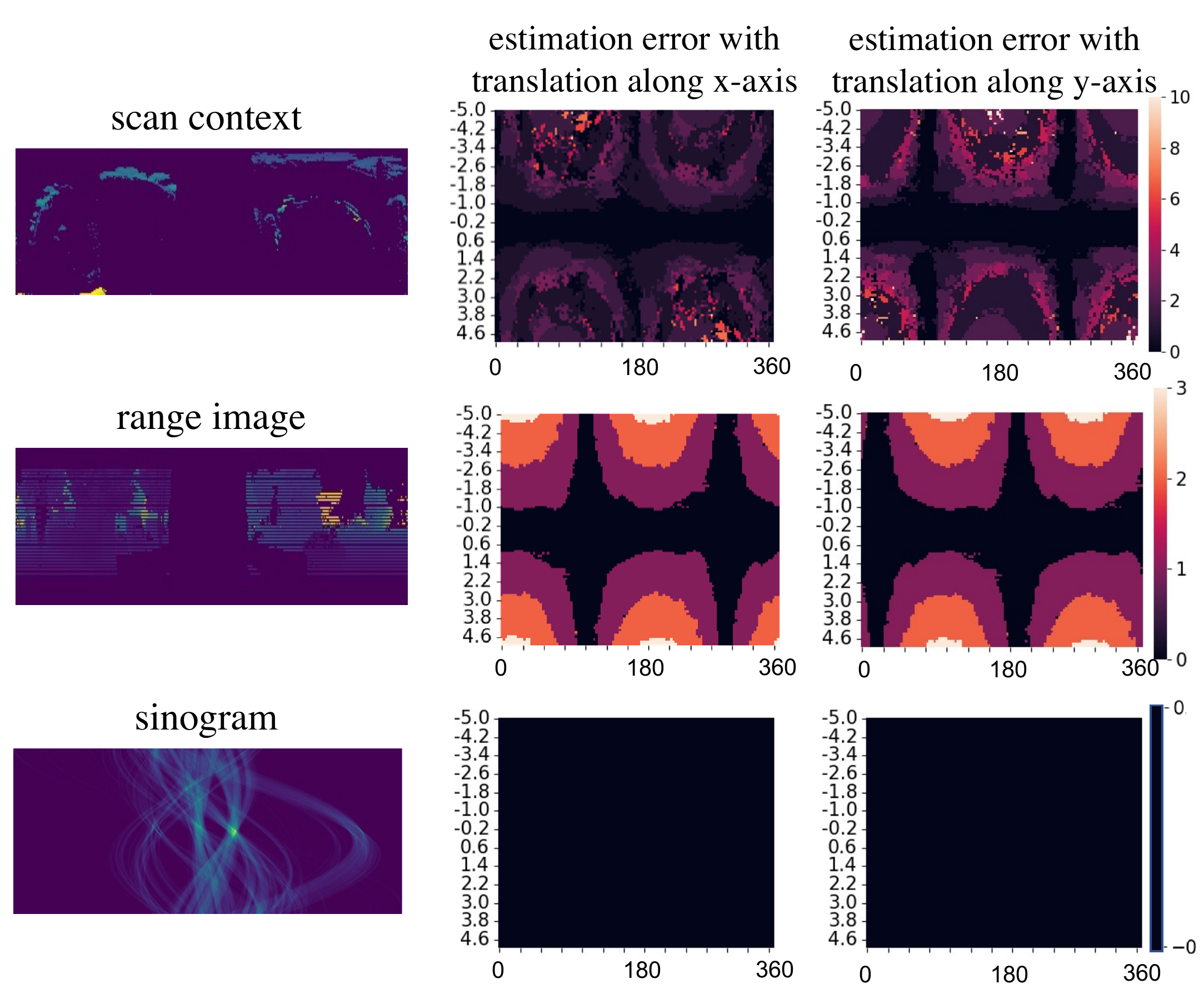}
    \caption{The toycase study that reflects the influence of relative translation between point clouds on heading angle estimation. The first column shows three representations and the second and third columns show the heading angle estimation errors computed based on these representations. The x-axis in the second and third columns denotes the rotated heading angles, and the y-axis denotes the relative translation.}
    \label{fig:trans_inv}
\end{figure}

% scan context, spherical (OREOS, overlapnet)
% translational invariance
% intermediate visualization
% To reveal the translation invariance of the proposed representation for heading angle estimation, we first design a toycase in which we transform a point cloud by translation between $[-5, 5]$ meters and rotation angles between $[0,360)$ degrees, then estimate the angles between the transformed and original point clouds using our representation and the other polar image based representations with the same circular cross-correlation backend. 
We first design a toycase to validate the translation invariance of the proposed method for heading angle estimation. Specifically, we translate a point cloud by a distance between $[-5m,5m]$  and rotate it by an angle between [0$\degree$, 360$\degree$). We then compare different representations by estimating the transformation between the original and the transformed point clouds, adopting the same circular cross-correlation backend.
The comparison representations include the polar BEV based Scan Context\cite{kim2018scan} and the range image used in \cite{oreos19,chen2020overlapnet}. The heading angle resolution in all the representations is 1$\degree/$pixel. The Scan Context is generated with the max range and ring number as $100$. The height of the range image is $100$ with $50\degree$ field of view. The representations and their estimation errors are drawn in Fig. \ref{fig:trans_inv}. 

From the results we can see that the translation distances largely influence the results estimated by Scan Context and range image representations. They can only achieve good performance when the translation is small. 
While the results estimated by our proposed representation are not affected. As the distributions of the two point clouds are theoretically the same, our method can achieve zero estimation error on all the relative transformations as shown in Fig. \ref{fig:trans_inv}. In real-world experiments, as there exist occlusions and noises, the distributions of point clouds would be different.  

\begin{figure}
    \centering
    \includegraphics[width=0.48\textwidth]{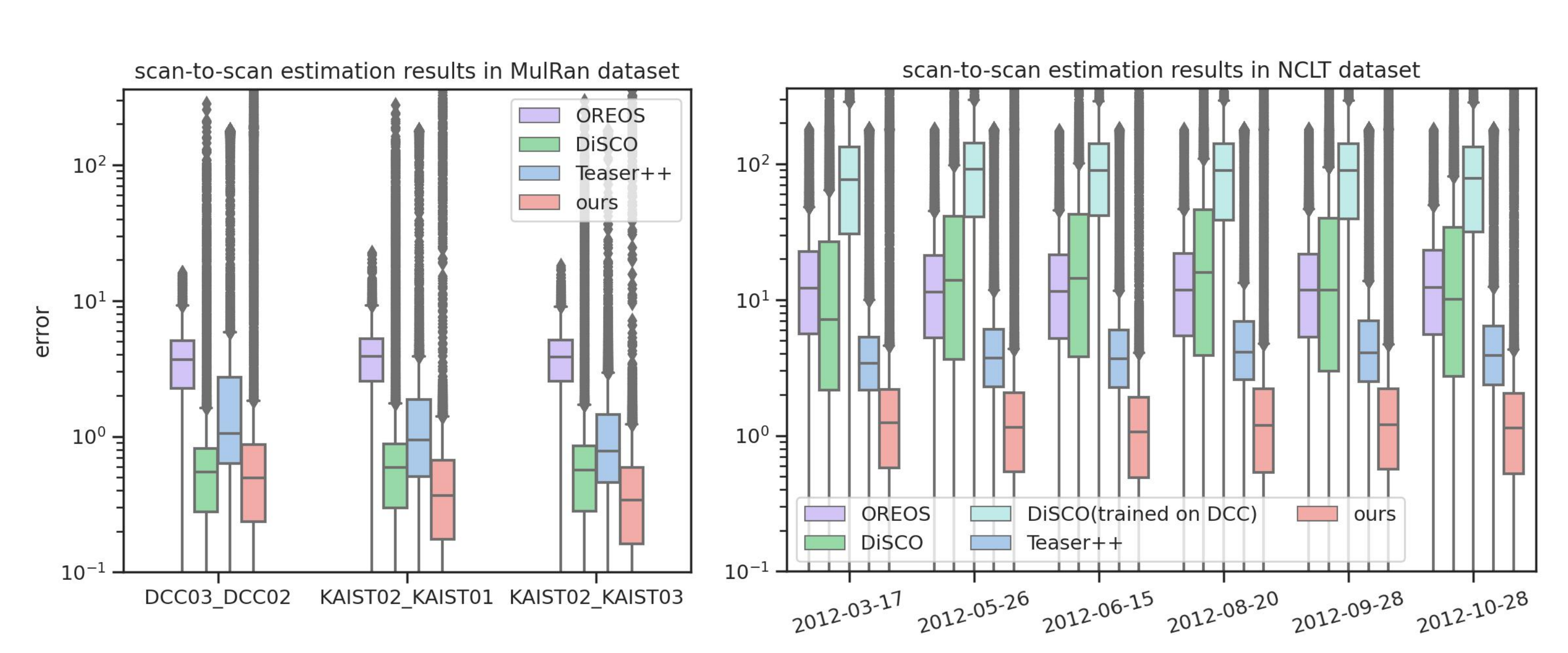}
    \caption{Scan-to-scan estimation results on MulRan and NCLT datasets.}
    \label{fig:scan_scan}
\end{figure}

% \subsection{Real-world Study}
% It is a promising property in applications that the proposed method can generalize across scenarios, robots and sensors. 
% datasets and settings
% scan to scan
% scan to submap
% translation invariance
% Time/storage
\subsection{Datasets and experimental setting }\label{datasetting}
We utilize MulRan\cite{kim2020mulran} and NCLT\cite{NCLTdataset} datasets for real-world experiment verification. The performance of heading angle estimation between laser scans and laser scan to laser submap are both evaluated. A laser submap is constructed using laser scans collected within 50 meters. For BEV based representations we remove the ground points following \cite{xu2021disco}.

\textbf{MulRan dataset} \cite{kim2020mulran} is a multimodal range dataset for urban place recognition. The data is collected in four places in Korea, with three trajectories collected in each place. We reserve the first trajectory in DCC (DCC01) for network training, and evaluate the performance on DCC02, DCC03 and the three trajectories collected in KAIST. Considering the overlaps across trajectories, we utilize DCC03 as the map and retrieve scans/submaps from it for each scan in DCC02. In KAIST dataset the trajectory KAIST02 is set as the map. Note that the scenario is changed from DCC to KAIST, thus the evaluation on KAIST can be regarded as testing generalization across places. The retrieval occurs if relative translation between two point clouds is within $5m$. 

\textbf{NCLT dataset} \cite{NCLTdataset} is a long-term vision and lidar dataset collected in the University of Michigan’s North Campus. A Velodyne HDL-32E 3D LiDAR is equipped on a Segway robot for data collection. We utilize the NCLT dataset to validate the generalization performance as both the scenario and sensors are changed. The trajectory collected on ``2012-01-08" is utilized as the map, and we test the heading angle estimation on the other six trajectories collected on ``2012-03-17", ``2012-05-26", ``2012-06-15", ``2012-08-20", ``2012-09-28" and ``2012-10-28". The other settings are kept the same with MulRan dataset. Note that for the compared methods that show poor generalization performance, we retrain their networks using the trajectories ``2012-02-04''.

\subsection{Compared methods} \label{compare_method}
We compare with learning-based methods OREOS \cite{oreos19} and DiSCO\cite{xu2021disco}, and one of the state-of-the-art 3D global point cloud registration method Teaser++\cite{TeaserTRO2021}. Note that our method is training-free when the point clouds have similar distributions. To get rid of the influence from the place recognition supervision, all the networks are retrained using the same data and only supervised by the rotation loss. The random circular column shift is applied during the network training process.

\textbf{OREOS}\cite{oreos19} is implemented using PyTorch following the network structure in their paper. We generate the input range images with the size of $32\times360$.

\textbf{DiSCO}\cite{xu2021disco} utilizes the multi-layer Scan Context as input. We generate the input into 20 layers following the settings in their paper with the rotational resolution of 1\degree per pixel. The size of the input is $40\times360\times20$. 

\textbf{Teaser++}\cite{TeaserTRO2021} can robustly register two point clouds in the presence of a large percentage of outlier correspondences. We utilize FPFH\cite{fpfh2009} feature descriptors for correspondence generation. The input point clouds are downsampled by a voxel filter with $0.5m$ resolution.

% \textbf{Ours} estimates the heading angle between generated BEV images with the size of $400\times 400$. 

\subsection{Scan-to-scan heading angle estimation}
We first perform the scan-to-scan heading angle estimation in MulRan and NCLT datasets by retrieving the original laser scans for estimation. Since the density distributions between the retrieved and query point clouds are similar, in our proposed method we do not utilize the network for feature extraction and directly estimate heading angles based on BEVs. While for OREOS\cite{oreos19} and DiSCO\cite{xu2021disco} the networks are trained on DCC01 and ``2012-02-04'' respectively for each dataset as noted in \ref{datasetting} and \ref{compare_method}. The results are drawn in Fig. \ref{fig:scan_scan}. We also summarize the trajectories in each dataset and list the statistics in TABLE \ref{tab:scan-scan}. The statistics include the percentages of estimation results that are less than the set thresholds $(1\degree/3\degree/5\degree)$, and the quartile encoding the estimation errors at the percentage of $25\%$, $50\%$ and $75\%$ of the whole results.

As the results show, our method could achieve the smallest estimation errors compared with the other methods with no need to extract features using network. Based on the results in the table we can see that the regression-based learning method OREOS shows larger estimation errors compared with the others that solve the results with structural constraints. And Teaser++ solves the relative transformation between point clouds in 6DoF, thus showing larger estimation errors than DiSCO in MulRan dataset in which the car drives on the planar road. However in NCLT dataset, the degree of freedom of the Segway robot includes the rotation in pitch, thus the noise of gravity alignment would affect the representations in OREOS and DiSCO, while Teaser++ could achieve better performance. Note that though the DiSCO also utilizes BEVs as input, the height information is discretized and used in the multi-layer representation, which increases the sensitivity to the gravity alignment and height differences between sensors. As only occupancy information is used in our representation, we could achieve better performance compared with the other methods in NCLT dataset. 

\begin{table}[htbp]
  \centering
  \caption{Scan-to-scan heading angle estimation results}
  \setlength{\tabcolsep}{0.9mm}{
    \begin{tabular}{ccccc}
    \toprule
    \multirow{2}[2]{*}{methods} & \multicolumn{2}{c}{MulRan} & \multicolumn{2}{c}{NCLT} \\
          & \multicolumn{1}{c}{$1\degree/3\degree/5\degree$}$\uparrow$ & \multicolumn{1}{c}{quartile($\degree$)}$\downarrow$ & \multicolumn{1}{c}{$1\degree/3\degree/5\degree$}$\uparrow$ & \multicolumn{1}{c}{quartile($\degree$)}$\downarrow$ \\
    \midrule
    OREOS &   0.08/0.34/0.73    &   2.46/3.80/5.15    &   0.05/0.14/0.23    & 5.39/11.84/21.90 \\
    DiSCO &   \textbf{0.87/0.95/0.96}    &   0.29/0.57/0.85    &  0.10/0.25/0.34     & 3.03/11.67/38.37 \\
    Teaser++ &   0.54/0.86/0.94    &    0.52/0.91/1.88   &   0.04/0.37/0.65    &2.35/3.79/6.21  \\
    Ours  &   0.86/\textbf{0.95/0.96}    &    \textbf{0.18/0.39/0.70}   &   \textbf{0.44/0.86/0.93}    &  \textbf{0.54/1.17/2.11}\\
    \bottomrule
    \end{tabular}%
    }
  \label{tab:scan-scan}%
\end{table}%

\subsection{Scan-to-submap heading angle estimation}
\begin{figure}
    \centering
    \includegraphics[width=0.48\textwidth]{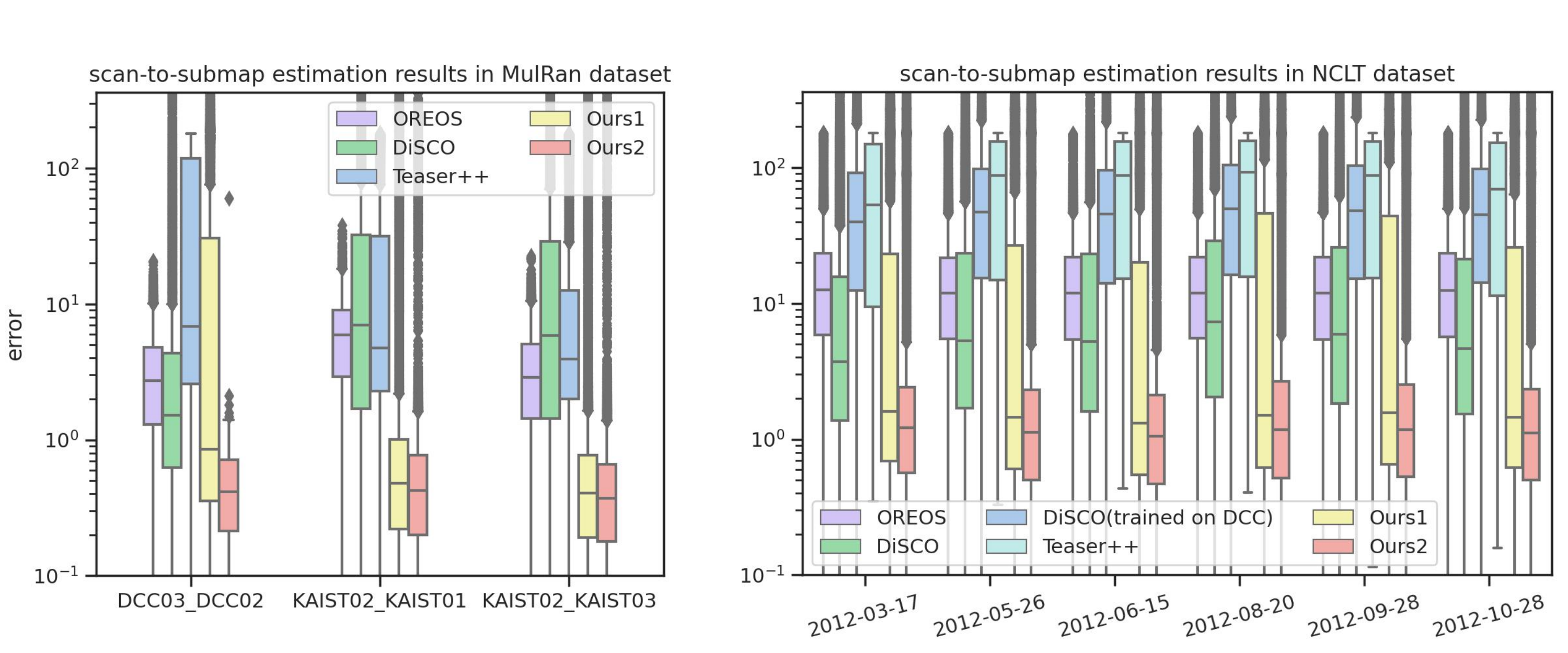}
    \caption{Scan-to-submap estimation results on MulRan and NCLT datasets. ``Ours1'' denotes our results evaluated without network, and ``Ours2'' denotes our results evaluated with network trained on DCC.}
    \label{fig:scan_submap}
\end{figure}

% Here we train our network based solution and the networks of the compared methods\cite{oreos19,xu2021disco} using scan-to-submap data. 
We evaluate the performance of heading angle estimation between the scan and submap point clouds. Note that the contents of BEVs generated by scan and submap are largely different, thus we utilize two feature extraction networks with the same structure to learn features respectively on scan and submap in our network and DiSCO. The results are shown in Fig. \ref{fig:scan_submap} and the statistics are listed in TABLE \ref{tab:scan_submap}.

\begin{table}[htbp]
  \centering
  \caption{Scan-to-submap heading angle estimation results}
  \setlength{\tabcolsep}{0.3mm}{
  \begin{threeparttable}
    \begin{tabular}{lcccc}
    \toprule
    \multirow{2}[2]{*}{methods} & \multicolumn{2}{c}{MulRan} & \multicolumn{2}{c}{NCLT} \\
          & \multicolumn{1}{c}{$1\degree/3\degree/5\degree$}$\uparrow$ & \multicolumn{1}{c}{quartile($\degree$)}$\downarrow$ & \multicolumn{1}{c}{$1\degree/3\degree/5\degree$}$\uparrow$ & \multicolumn{1}{c}{quartile($\degree$)}$\downarrow$ \\
    \midrule
    OREOS &   0.15/0.45/0.65    &   1.65/3.43/6.37    &   0.05/0.14/0.23    & 5.56/12.09/22.35 \\
    DiSCO &   0.25/0.47/0.56    &   1.01/3.51/20.21    &  0.16/0.38/0.49     & 1.65/5.16/22.86 \\
    Teaser++ &   0.06/0.35/0.51    & 2.23/4.75/43.10   &   0.00/0.03/0.08    &12.99/80.04/154.20 \\
    Ours1 & {0.71/0.81/0.82} & {
    0.24/0.52/1.23}& {0.37/0.67/0.70} & {0.63/1.49/30.47} \\
    Ours2  & {\textbf{0.83/0.94/0.95}} & {\textbf{0.21/0.44/0.79}} & {\textbf{0.45/0.79/0.83}}   & {\textbf{0.52/1.15/2.39}}\\
    \bottomrule
    \end{tabular}%
    \begin{tablenotes}
    \footnotesize
    \item ``Ours1'' denotes our results evaluated without network.
    \item ``Ours2'' denotes our results evaluated with network trained on DCC.
    \end{tablenotes}
    \end{threeparttable}
    
    }
  \label{tab:scan_submap}%
\end{table}%

Based on the results we can notice that as the data distributions are largely different in BEV-based representation, the accuracy of ours and DiSCO's are decreased. But our performance is still better than the compared methods and 95\% of the estimation errors are less than 5\degree in MulRan dataset. As we utilize the scans with graph optimized poses for submap generation in NCLT dataset, the noise of gravity alignment is less compared with the one in the scan-to-scan test, thus the performance of DiSCO in NCLT dataset is better. The difference in data distributions also influences Teaser++ as the point cloud features cannot keep consistency in this situation. Also it is interesting to notice that, our method without a feature extraction network could also demonstrate good performance in many trajectories as the circular structural constraint is applied for solution exploration. And the feature extraction network helps improve the consistency between scan and submap, and can generalize even when both the scenarios and equipment are changed.

\subsection{Runtime Evaluation}
We evaluate the time cost of our method on a GeForce RTX 2070 GPU with an Intel i7-9700 CPU. The details of the time cost are listed in TABLE \ref{tab:time}. The ``Preprocess'' module projects the original 3D laser point clouds to generate the occupancy-based BEV images. The ``Feature extraction'' module calculates the translation invariant representations from original or masked BEVs. ``Heading angle estimation'' performs the circular cross-correlation to calculate the global optimal results. Based on the results we can find that if the point clouds are in the same distribution and the network is not required for feature extraction, only around $1ms$ is taken to compute the global heading angle. If the network is used the total time cost is around $53ms$, which also meets the requirement for real-time computation.

\begin{table}[htbp]
  \centering
  \caption{Computational cost of the proposed method}
  \setlength{\tabcolsep}{0.5mm}{
    \begin{tabular}{ccccc}
    \toprule
          & Preprocess    &  Feature extraction&  Heading angle  & Total \\
          & & (wo/w network) & estimation&(wo/w network) \\
    \midrule
    time(ms) & 0.58      &  0.28/52.1     &   0.21    &  1.07/52.89\\
    \bottomrule
    \end{tabular}%
    }
  \label{tab:time}%
\end{table}%

\section{Conclusion}
In this paper, a translation invariant representation is proposed for heading angle estimation between gravity-aligned point clouds. The sinogram is generated to transform the occupancy-based BEV into the polar image based on Radon Transform. We further eliminate the influence of translation on this representation, and achieve global optimal heading angle estimation efficiently using circular cross-correlation. To address the point clouds in different distributions, we train a feature extraction network with the differentiable heading angle estimator integrated. And the trained model can be generalized to different places and sensors. The experimental results validate both the accuracy and efficiency of the proposed method compared with the other methods.

In the future, we propose to integrate this method into the full localization system, and focus on heading angle estimation on heterogenous maps.

\addtolength{\textheight}{-2cm}   % This command serves to balance the column lengths

%%%%%%%%%%%%%%%%%%%%%%%%%%%%%%%%%%%%%%%%%%%%%%%%%%%%%%%%%%%%%%%%%%%%%%%%%%%%%%%%

%%%%%%%%%%%%%%%%%%%%%%%%%%%%%%%%%%%%%%%%%%%%%%%%%%%%%%%%%%%%%%%%%%%%%%%%%%%%%%%%


% \section*{ACKNOWLEDGMENT}

%%%%%%%%%%%%%%%%%%%%%%%%%%%%%%%%%%%%%%%%%%%%%%%%%%%%%%%%%%%%%%%%%%%%%%%%%%%%%%%%

\bibliographystyle{IEEEtran}
\bibliography{mybib}

\end{document}